\documentclass[10pt, conference, letterpaper]{IEEEtran}

\IEEEoverridecommandlockouts
\usepackage{cite}
\usepackage{amsmath,amssymb,amsfonts}
\usepackage{algorithmic}
\usepackage{graphicx}
\usepackage{textcomp}
\usepackage{xcolor}
\usepackage{booktabs}
\usepackage{diagbox}
\usepackage{multirow}
\usepackage{subfigure}
\usepackage{caption}
\usepackage{enumitem}
\usepackage{pifont}

\def\BibTeX{{\rm B\kern-.05em{\sc i\kern-.025em b}\kern-.08em
    T\kern-.1667em\lower.7ex\hbox{E}\kern-.125emX}}

\newcommand{\tb}{\noindent\textbf}
\newcommand{\eb}{\vspace{0.5em}\noindent\emph}

\newcommand{\ie}{\emph{i.e.}}
\newcommand{\eg}{\emph{e.g.}}

\begin{document}


\title{AccDecoder: Accelerated Decoding for Neural-enhanced Video Analytics}



\author{\IEEEauthorblockN{Tingting Yuan\IEEEauthorrefmark{2}\IEEEauthorrefmark{4},
Liang Mi\IEEEauthorrefmark{3}\IEEEauthorrefmark{4}, Weijun Wang\IEEEauthorrefmark{2}\IEEEauthorrefmark{3},
Haipeng Dai\IEEEauthorrefmark{3}\IEEEauthorrefmark{1}, 
Xiaoming Fu\IEEEauthorrefmark{2}}
\thanks{*Corresponding author. §Equal contributors.}

\IEEEauthorblockA{
\IEEEauthorrefmark{2}University of G{\"o}ttingen, Germany; 
\IEEEauthorrefmark{3}Nanjing University, China
\\
\{tingting.yuan,weijun.wang,fu\}@cs.uni-goettingen.de,
liangmi@smail.nju.edu.cn,
haipengdai@nju.edu.cn}}

\maketitle

\begin{abstract}
The quality of the video stream is key to neural network-based video analytics.
However, low-quality video is inevitably collected by existing surveillance systems because of poor quality cameras or over-compressed/pruned video streaming protocols, \eg, as a result of upstream bandwidth limit.
To address this issue, existing studies use quality enhancers (\eg, neural super-resolution) to improve the quality of videos (\eg, resolution) and eventually ensure inference accuracy.
Nevertheless, directly applying quality enhancers does not work in practice because it will introduce unacceptable latency.
In this paper, we present AccDecoder, a novel accelerated decoder for real-time and neural-enhanced video analytics.
AccDecoder can select a few frames adaptively via Deep Reinforcement Learning (DRL) to enhance the quality by neural super-resolution and then up-scale the unselected frames that reference them, which leads to 6-21\% accuracy improvement.
AccDecoder provides efficient inference capability via filtering important frames using DRL for DNN-based inference and reusing the results for the other frames via extracting the reference relationship among frames and blocks, which results in a latency reduction of 20-80\% than baselines.


\end{abstract}


\section{Introduction}


Advances in computer vision offer tremendous opportunities for autonomous analytics of videos generated by pervasive video cameras.
Deep Neural Networks (DNNs) \cite{bochkovskiy2020yolov4, ahn2018fast, ren2015faster, wang2021swiftnet} have been developed to dramatically improve the accuracy for various vision tasks, while introducing stringent demands on computational resources.
Due to the compute-resource shortage of commercial cameras, videos need to be streamed to powerful servers for inference, which is called distributed video analytics pipeline (VAP) \cite{du2020server, li2020reducto}.

Nevertheless, providing highly accurate video analytics remains challenging for the state-of-the-art distributed VAPs.
Since most methods for video analytics currently rely on high-resolution videos, it is difficult to analyze low-quality videos, such as object detection at low resolutions.
For example, the accuracy of Faster R-CNN \cite{ren2015faster}, a modern DNN-based inference method, can only achieve around 56\% accuracy for videos in 360p and 61\% accuracy for videos in 540p which are collected by \cite{Yoda_c}.
However, low-quality videos are inevitably collected by existing surveillance systems.
One of the reasons is that existing low-quality collectors can only capture low-resolution frames. 
For example, New York city’s department of transportation \cite{cameras} has made videos from all 752 traffic cameras to the public; however, the videos are transmitted at an extremely low resolution (240p) due to the default configuration of cameras \cite{wei2019city}.
Another reason is that current video streaming protocols over-compress/prune videos due to upstream bandwidth limitations.
For example, AWStream \cite{Awstream} aggressively reduces the resolution of the video from 540p to 360p and the frame rate from 1 to 0.83.
It eventually brings about a 66\% saving of bandwidth with a reduced accuracy from 61\% to 54\%.

To address this challenge, some VAPs \cite{yi2020eagleeye, yi2020supremo} try to utilize image enhancement models like Super Resolution (SR) \cite{ahn2018fast,huang2017speed} and Generative Adversarial Network (GAN) \cite{chen2018fsrnet} to enhance frames in videos before feeding them into the inference model.
This idea is inspired by the observation from the computer vision community -- running object recognition-related tasks on high-resolution images can largely improve the detection accuracy \cite{huang2017speed}.
However, the video enhancement via DNN-aware image enhancement models introduces extra latency, resulting in around 500 ms end-to-end latency \cite{wang2022enabling} for each frame, which is far from the real-time requirement (\eg, less than 15-30 ms for real-time object recognition \cite{li2020reducto}).

Although existing DNN-aware video enhancement provides a promising way to improve the inference accuracy \cite{zou2019object}, there is still much room for improvement.
First, prior video enhancement mechanisms are largely agnostic to video contents, treating each received frame equally, but not all the frames need to be enhanced.
For example, only the frames containing vehicles are valuable for traffic flow analysis; on the contrary, enhancing frames with empty streets is worthless but only increases system latency.
Therefore, content-agnostic enhancement mechanisms cannot avoid being suboptimal.
Second, although new DNN frameworks are designed to accurately recognize important frames (\eg, \cite{ghodrati2021frameexit, habibian2021skip}), they are too heavy to achieve low latency.
Third, decoding all the frames for analytics is computationally intensive and time-consuming, and video encoding contains plenty of unexploited but handy information to capture the important frames, such as motion vectors (MVs) and residuals.
We argue that the codec information, although inaccurate, is valuable to reveal which content is important, thereby speeding up video analytics.

Motivated by the above insights, we present AccDecoder, a novel accelerated video streaming decoder for real-time and neural-enhanced video analytics.
AccDecoder is a content-aware DNN-integrated video decoder utilizing codec information to select a few frames for quality enhancement, which is called as \textit{anchor frames} and some frames for DNN-aware inference, which is called as \textit{inference frames}.
 In particular, AccDecoder applies SR to anchor frames and transfers the high quality of frames/blocks to benefit the entire video via extracting the frame and block reference relationships; it further leverages codec information to reuse the results of inference frames for acceleration.

\begin{figure}[t!]
\centering
\hspace{-0.5em}
\includegraphics[width=0.26\textwidth]{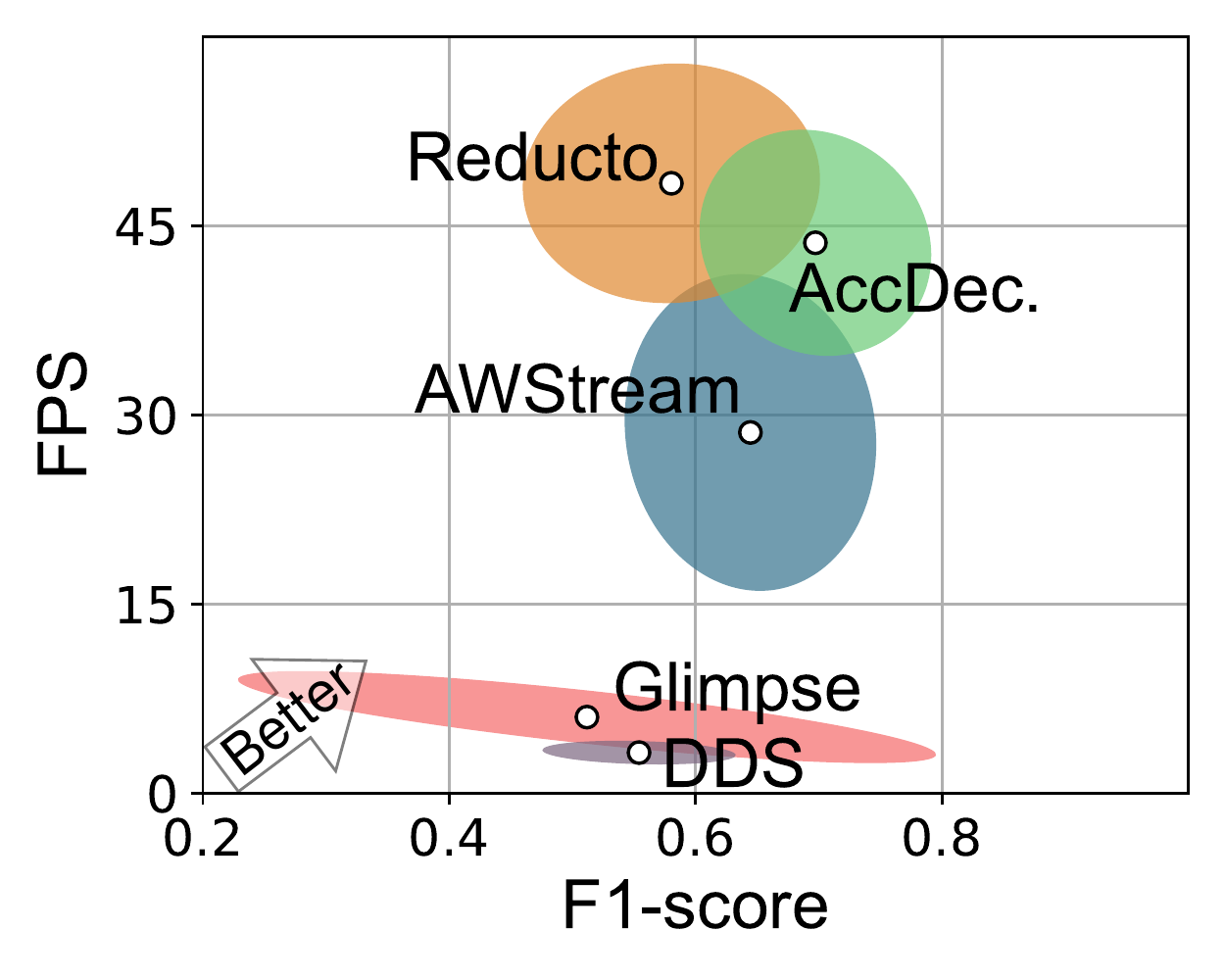}
\caption{Example results of AccDecoder vs. baselines.}
\label{fig:s_bub}
\vspace{-1.5em}
\end{figure}

\vspace{0.5em}
\tb{Challenge and solution.}
AccDecoder needs to adapt to various quality of videos to enable robust and real-time video analytics.
Since SR and DNN-based inference is time-consuming, an accuracy-latency tradeoff must be made carefully to keep low latency without drastically compromising the accuracy.
Our preliminary study (see more in Section II and III) shows that AccDecoder needs adaptive metrics for anchor and inference frame selection according to factors like video content and quality.
In other words, various videos (or even different chunks of the same video) demand different settings for frame selection, but a static setting is not adaptive to the varying contents of videos.
To address this issue, we leverage deep reinforcement learning (DRL) \cite{luong2019applications,arulkumaran2017brief, yuan2022dacom}, which has been widely used to solve dynamic and sequential problems \cite{yuan2021harnessing,yuan2020dynamic}.
Specifically, AccDecoder enables adaptive settings via DRL in frame selection to accelerate the video analytics and achieve a good tradeoff between accuracy and latency.



\noindent
\textbf{Our contributions} are summarized as follows.
\begin{itemize}[left=0.2em]
\item We design a novel content-aware DNN integrated video decoder that is resilient and robust to the quality of videos.
For example, for a 540p crossroad video collected by \cite{Yoda_c}, AccDecoder can improve 10-38\% accuracy compared with the state-of-the-art VAPs (see Fig. \ref{fig:s_bub}).
\item We exploit temporal redundancies within a video and codec information to drastically increase the speed of analytics.
For example, AccDecoder performs 1.5-8.0 times faster compared with AWStream \cite{Awstream}, DDS \cite{du2020server}, and Glimpse \cite{chen2015glimpse} (see Fig. \ref{fig:s_bub}).
\end{itemize}

The rest of the paper is organized as follows.
We first present the background and motivation in Section II.
Then we discuss AccDecoder's key design in Section III, followed by our implementation in Section IV. The experimental studies are demonstrated in Section V. In Section VI, we introduce some related work. We conclude this paper in Section VII.

\section{Background and Motivation}

We introduce distributed VAPs and video codec, followed by performance requirements of VAPs that drive our design (\ie, high accuracy, low end-to-end latency, and low overheads).
We then elaborate on why prior solutions struggle to meet the three requirements simultaneously.

\subsection{Background}
\begin{figure}[h!]
\centering
    \includegraphics[width=0.49\textwidth]{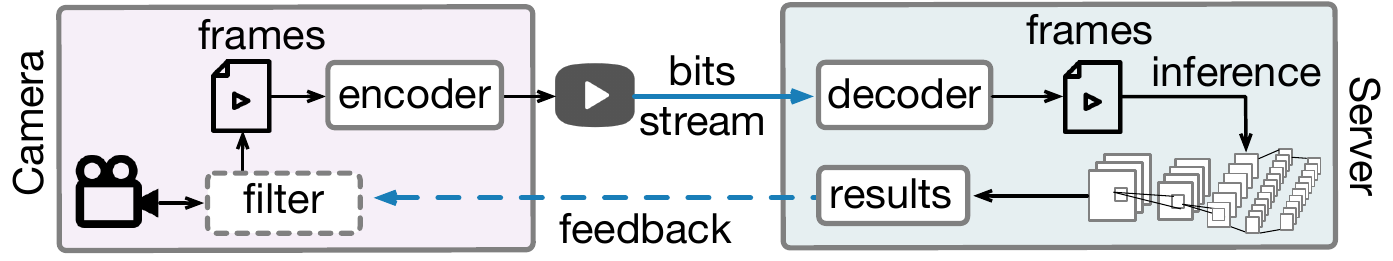}
    \caption{Distributed video analytics pipeline.}
    \label{fig:pip}
\end{figure}

\tb{Distributed Video Analytics.}
The proliferation of video analytics is facilitated by the advances of deep learning and the low prices of high-resolution network-connected cameras.
However, the accuracy improvement from deep learning comes at a high computational cost.
Although state-of-the-art smart cameras can support deep learning methods, the current surveillance and traffic cameras show only suboptimal use of resources.
For example, DNNCam \cite{DNNCam} that ships with a high-end embedded NVIDIA TX2 GPU \cite{JETSONTX2} costs more than \$2000 while the price of deployed traffic cameras today ranges \$40-\$200.
These cameras are typically loaded with a single-core CPU, only providing scarce compute resources.
Because of this huge gap, typical VAPs follow a distributed architecture.
A typical distributed VAP architecture includes a filter and an encoder on the camera side and a decoder and an inference model on the server side, as illustrated in Fig. \ref{fig:pip}.
In live analytics, video frames are continuously encoded and sent to a remote server that runs the inference DNN to analyze the video in an online fashion.
For example, a vehicle detection pipeline consists of a front-end traffic camera (which compresses and streams live videos to a compute-powerful edge/cloud GPU server upon wire/wireless networks) and a back-end server (which decodes received video into frames and feeds them into inference models like Faster R-CNN \cite{ren2015faster} to detect vehicles).
Such an architecture brings challenges in bandwidth cost, and thus some schemes rely on aggressively pruning videos (via \eg, reconfiguration, filtering) to meet upstream bandwidth limitations.

\tb{Video codec}\label{codec} is the key technology in distributed VAPs.
It comprises an encoder and a decoder, a software/hardware program used to compress/decompress video files for easier storage or network delivery.
The encoder compresses video data and wraps them into common video formats (\eg, H.264 \cite{wiegand2003overview}), while the decoder decompresses the compressed video data into frames before post-processing (\eg, playback or analysis). 
This compression process is usually lossy, which strikes a balance between the video quality and the compression ratio according to the self-preference of codecs and encoding settings of users (\eg, bitrate, frame rate, and group of pictures).
We take H.264, one of the most popular codecs, as an example to explain the compression process. 
During encoding, each video frame is ﬁrst divided into non-overlapped macroblocks (16×16 pixels), then to each macroblock, the encoder searches for the optimal compression method (including the block division types and encoding types of each block) according to the pixel-level similarity and encoding settings. 
One macroblock may be further divided into non-overlapped blocks (8×8 or 8×16 pixels) encoded with intra- or inter-frame types.
The intra-coded block is encoded using the reference block with the most pixel-value similarity, and the offset between these two blocks is encoded into an MV with a residual.
With the same procedure, the inter-coded block locates the most pixel-value similar block searched by the reference index and the MV from other frames.
An MV indicates the spatial offset between the target block and its reference, while the difference in pixel values of two blocks is encoded as the residual for decoding.

An ideal distributed VAP should meet three goals:
\begin{itemize}[left=0pt]
    \item \textbf{High accuracy}. Inspired by the success of image enhancement methods in video streaming for QoE improvement \cite{dasari2020streaming, kim2020neural,yeo2020nemo, yeo2018neural}, researchers try to use image enhancement for machine-centric video analytics \cite{wang2022enabling, wang2019bridging,yeo2018neural}. For example, \cite{wang2022enabling} leverages SR to enhance image details for robotics applications, while \cite{yeo2018neural} embeds GAN on Google Glasses to generate facial features for face recognition. The experimental results from \cite{wang2022enabling,yeo2018neural} demonstrate that image enhancement improves inference accuracy.
    
    
    \item \textbf{Low latency}. The latency of decoding the video (\ie, decoding latency), inference, and streaming the video to the server (\ie, streaming latency) should be low.
    Previous works focus on reducing the size of streaming to reduce the streaming latency (\eg, Reducto \cite{ren2015faster}), learning and filtering key frames for inference, and utilizing MVs to reuse the other frames.
    \item \textbf{Low bandwidth cost}. We define the total size of a video file delivered from the camera to the server of each VAP as its bandwidth cost. 

\end{itemize}

\subsection{Motivation}
\tb{Limitations of previous work.}
Although a couple of bandwidth-saving and accuracy-improvement approaches have been developed, three significant limitations remain.

\begin{itemize}[left=0pt]
   \item  \textbf{Adaptive encoding in cameras}.
    A camera may leverage light-weight DNN \cite{zhang2015design} or heuristic methods (\eg, inter-frame pixel-level difference \cite{chen2015glimpse}) to distinguish and prune frames/regions without labelled information. However, these cheap methods may cause false positive (\eg, pixel-level distance changes by background may trigger a camera to send many frames) and false negative (\eg,  cheap object detection model may miss small appeared objects), thus increasing bandwidth cost or reducing the inference accuracy. Server-side decision-making controls the camera's actions with feedback from the cloud.
    For example, according to the servers' instruction, the camera in \cite{du2020server} iteratively delivers the region of interest in a higher resolution. The server-side decision-making introduces extra latency, especially due to the information delivery between the camera and the cloud crossing a wide area network.
    
    
    \item \textbf{Image enhancement in servers}.
    Image enhancement contributes to higher accuracy but also causes higher latency. Although recent studies have successfully leveraged image enhancement to increase QoE and inference accuracy, they still suffer from high latency. The root cause is that image enhancement models are much heavier than other computer vision models. For instance, the complexity of the SR model is as high as 1000× heavier than image classification, and object detection models in terms of MultAdds \cite{yi2020supremo}, as SR outputs high-resolution (HR) images whereas others output labels or Bounding boxes (Bbox). In other words, naively enhancing each frame is not practical in real-time video analytics applications.
    
\end{itemize}

\vspace{-1em}
\begin{figure}[htp!]
\centering
    \includegraphics[width=0.37\textwidth]{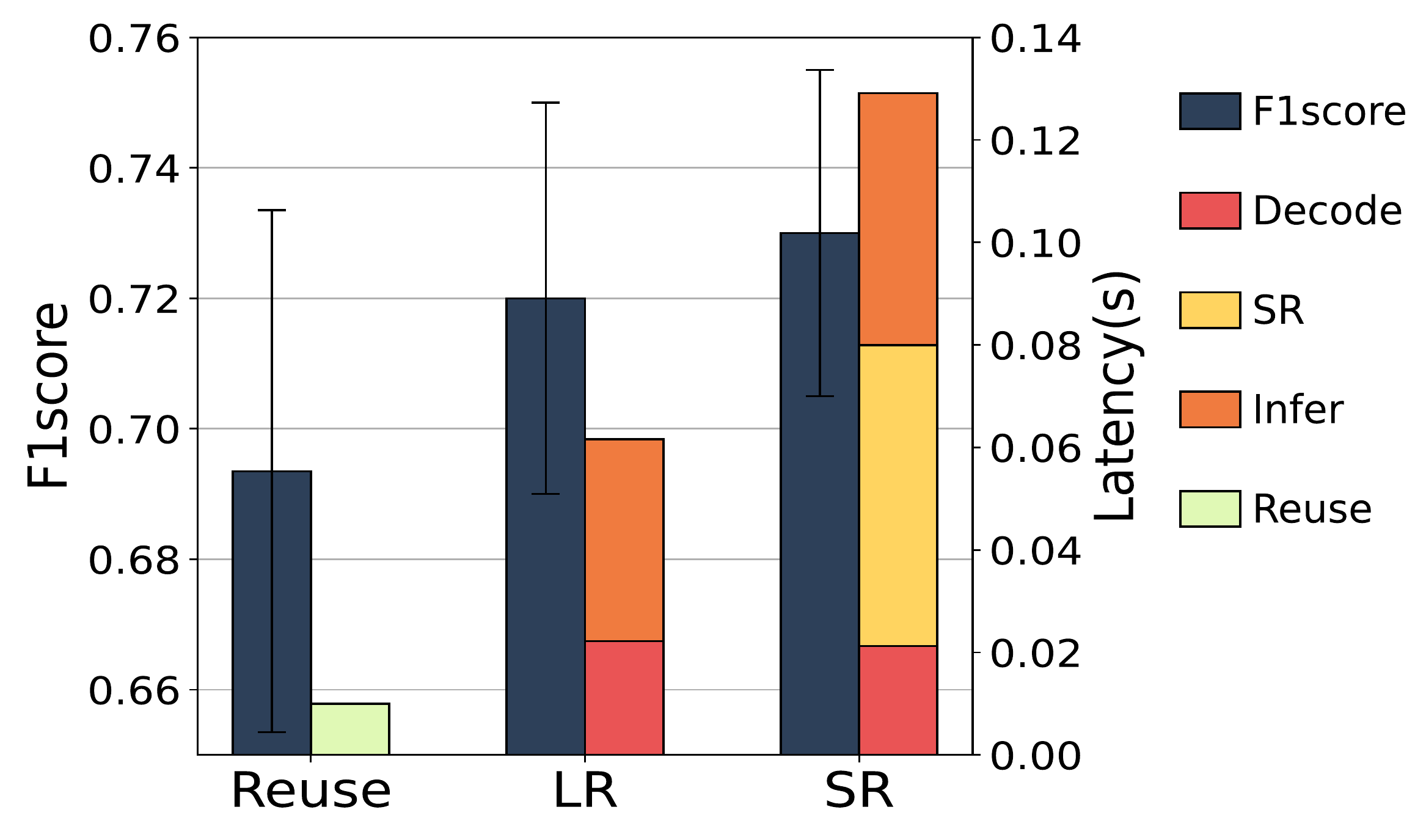}
    \vspace{-0.5em}
    \caption{Accuracy and latency using different schemes.}
    \label{fig:moti}
    \vspace{-0.5em}
\end{figure}

As shown in Fig. \ref{fig:moti}, our preliminary study confirms the challenge in the tradeoff between accuracy and latency in video analytics.
Due to resource restrictions, to use existing techniques in real-time and provide high accuracy, servers need to adaptively shift multiple pipelines, including inference after SR using HR frames, inference with low-resolution frames (LR), and reuse the last inference results.
However, conventional algorithms (\eg, k-nearest neighbor (KNN) used in \cite{li2020reducto}) are largely suboptimal as they ignore the temporal relationship among frames and the possibility of transferring high-quality frames to the whole video.
Therefore, we aim to design an efficient decoder to schedule the multiple pipelines for video analytics adaptively.




\section{System Design}
In this section, we present the design goals and our solution details of AccDecoder.

\begin{figure}[h!]
\centering
    \includegraphics[width=0.48\textwidth]{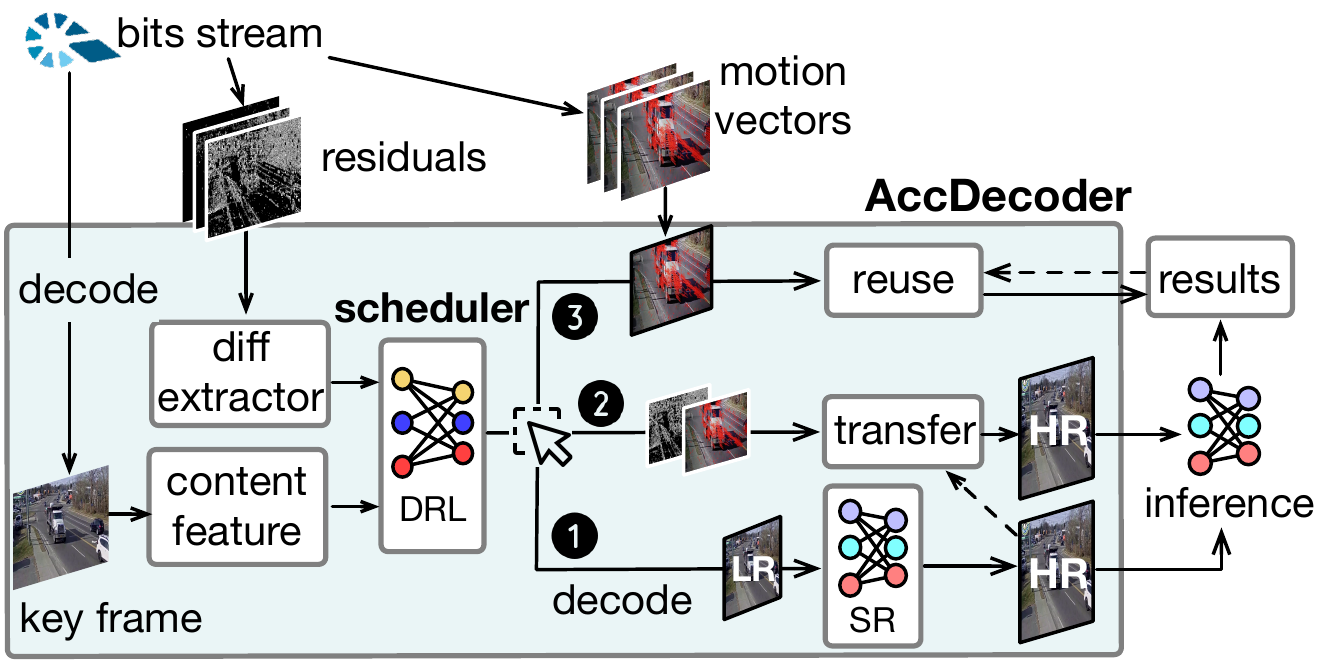}
    \caption{Architecture of AccDecoder.}
    \label{fig:AccDecoder}
\end{figure}

\tb{Design goals.} We take a pragmatic stance to focus on the server-side decoder because most of the installed surveillance or traffic cameras only have cheap CPUs without programmable ability~\cite{li2020reducto}; besides, rich information that may improve VAPs’ performance in the decoder has not been excavated.
In this context, we propose AccDecoder, a portable tool/decoder which can be plugged into any VAPs for video streaming analytics to achieve high accuracy, limited latency (\eg, 30 ms), and low-resource goals simultaneously.
As illustrated in Fig. \ref{fig:AccDecoder}, AccDecoder achieves these goals via the following three mechanisms:
1) Pipeline \ding{182} leverages \emph{SR} model enhancing a small set of LR \emph{anchor frames} to HR ones to achieve high accuracy.
2) Pipeline \ding{183} and Pipeline \ding{184} extract the codec information (\eg, frames reference relationship, MVs, and residuals) from the decoder, then utilize them to \emph{transfer} the gains of enhanced anchor frames and \emph{reuse} DNN inference results (\eg, Bbox in object detection) onto the entire video respectively.
\emph{Transfer} and \emph{reuse} amortize the computational overhead of SR and inference across the entire video and thus achieve low latency.
3) The \emph{scheduler} classifies all frames into three subsets, and each executes one of three pipelines.
It can greatly reduce latency and computational cost by exploiting the content features of the key frames (\eg, the intra-coded frame) and the change of codec information (\eg, residuals of continuous frames).

\tb{Path towards the goal.} We seek answers to the following pivotal questions, which lead to our key design choices.
Q1 --- How to effectively transfer the gains of SR to up-scale non-anchor frames?
Q2 --- How to reuse the results of inference to the other frames?
Q3 --- How to assign appropriate decoding pipelines to frames in a fine spatial granularity to achieve a better accuracy-latency tradeoff than baselines?

\eb{Q1 -- How to transfer gains of SR to non-anchor frames?}

\begin{figure}[htp!]
\centering
    \includegraphics[width=0.48\textwidth]{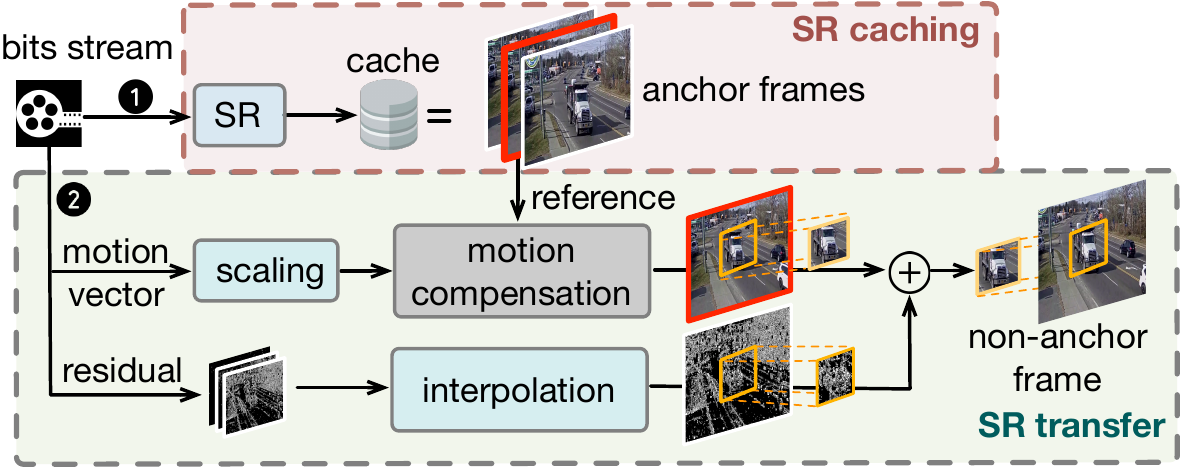}
    \caption{Process of SR gain transfer inspired by \cite{yeo2020nemo} and \cite{wiegand2003overview}.}
    \label{fig:IIIA}
\end{figure}

\tb{Approach: Pipeline \ding{182} + \ding{183}.} To make the best of reuse, AccDecoder enhances \emph{anchor frames} with the SR model and caches the output (see \ding{182} in Fig. \ref{fig:IIIA}); then it transfers the enhancement benefit to non-anchor frames with the reference information and the cached outputs (see \ding{183} in Fig. \ref{fig:IIIA}).
This approach follows the same findings in~\cite{ahn2018fast} and \cite{yeo2020nemo}, where most of the latency of SR occurs at the last couple of layers.
Namely, caching and reusing the final output (\ie, high-resolution images) is most effective in achieving low latency.

Fig. \ref{fig:IIIA} illustrates the process of transferring SR gains to a non-anchored frame for the inter-coded type.
Modern video codec encodes/decodes frames on the basis of non-overlapped \emph{inter-} and \emph{intra-coded} blocks (\S \ref{codec}).
AccDecoder uses the reference index, MVs, and the residual in the codec information to decode a target block.
The process is the same as normal decoding except for the additional SR, scaling, and interpolation modules (blue boxes in Fig. \ref{fig:IIIA}).
First, AccDecoder selects the reference blocks among cached anchor frames following the inference index.
Next, AccDecoder up-scales the MV with the same amplification factor as SR (\eg, from 270p to 1080 is 4).
Following the MV, AccDecoder transfers the SR gain from the reference block in the cached frame to the target one.
At last, AccDecoder up-scales the residual by light-weight interpolation (\eg, bilinear or bicubic), accumulates it to the transferred block to output the HR block, and pastes on the non-anchor frame.
To the intra-coded blocks without the cached reference anchor frame, AccDecoder directly decodes and up-scales then by interpolation.
Fortunately, with our carefully designed scheduler, most intra-coded blocks are assigned to the SR pipeline; the impact of the minority of interpolated intra-coded blocks can be negligible.

\eb{Q2 --- How to reuse inference results for non-inference frames?}

\begin{figure}[htp!]
\centering
    \includegraphics[width=0.46\textwidth]{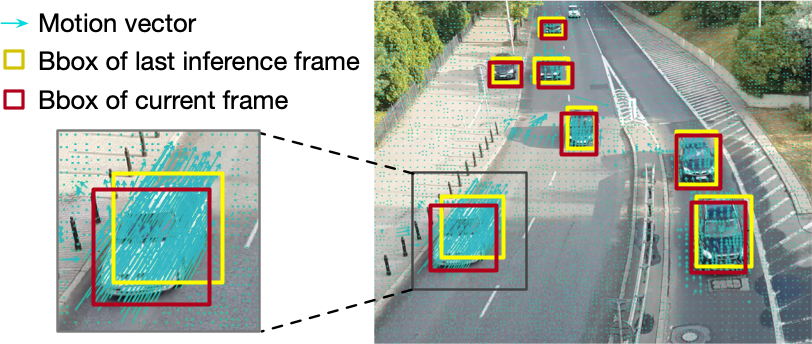}
    \caption{Relation between MVs and Bboxes.}
    \label{fig:mv}
\end{figure}

\tb{Approach: Pipeline \ding{184}.}
AccDecoder infers \emph{inference frames} using the inference model and caches the results; then, it uses the MVs and cached results to infer non-inference frames (see \ding{184} in Fig. \ref{fig:AccDecoder}). 
MV indicates the offset between the target and reference blocks (\S \ref{codec}).
Here we use the object detection task as an example, which aims to identify objects (\ie, their locations and classes) on each frame in videos. 
Fig. \ref{fig:mv} reveals that the MVs between the last inference frame and the current frame can perfect match the movement of objects’ Bboxes (\ie, the results of object detection).


The \emph{Reuse} module in Pipeline \ding{184} gets the result of the last inference frame (dotted line in Fig. \ref{fig:AccDecoder}), calculates the mean of all MVs that reside in each Bbox, and uses it to shift each Bbox to the current position.
MV, as the block-level offset, is hard to express any semantic meaning (\eg, object moving) \cite{yeo2017will}. 
Our preliminary study implies that the accuracy of \emph{reuse} degrades significantly after the MV, which spans multiple reference frames (\eg, over 7-10 frames in \cite{Yoda_c, Gebhardtdata, Yoda}).

\tb{Optimization.}
AccDecoder uses two techniques to improve the accuracy of the reuse of inference results.
First, it filters the noisy MVs from static backgrounds and outliers. 
Empirically, AccDecoder filters the MV whose value is equal to zero or greater than the mean plus 0.8 times the standard deviation in the Bbox to which it belongs.
Second, to cope with the change in Bbox size due to the object's movement, AccDecoder expands the MV calculation region to each direction by one macroblock (16 pixels).
Note that the reuse module is not necessary to tackle but only remit this erosion because the \emph{scheduler} module in (\emph{Q3}) effectively controls it by judiciously distributing inference frames.

\tb{Progress beyond the state of the art}.
Some prior work (\eg, \cite{zhang2020mobipose, xu2020approxdet, liu2019edge}) leverages lightweight MV-based methods to reuse analytics results and speed up inference.
However, rather than calculating MV between continuous frames in the playback order like them, reuse in AccDecoder works in compressed-video space.
That is, the reference blocks used to calculate the target block's MV can be distributed throughout the video (\emph{the target block can even refer to future frames under the playback sequence, which is called backward reference})\footnote{Modern video codecs aim at reducing video size; they only consider how to reduce the volume without caring whether the encoding obeys the playback order; thus \emph{coding order} is very different from the playback order.}, but the inference results should output in the playback order.
To tackle this mismatch, we maintain a graph to map the coding order to the playback order and accumulate the MVs along the edges.
Via statistical experiments on large-scale datasets, we find that the number of forwarding and backward references is less than 4 and 3 frames, respectively, so it is not time-consuming to search and calculate MVs in the graph.

\eb{Q3 --- How to guarantee accuracy-latency balance?}
\label{features}

\tb{Approach: Scheduler.} The key to the accuracy-latency tradeoff is how to optimally assign decoding pipelines (\ie, SR, inference, and reuse) to frames in fine spatial granularity.
As analyzed in Fig. \ref{fig:moti}, different pipelines for frame decoding and analytics lead to different levels of accuracy and latency.
For each frame in an SR class, selected anchor frames are enhanced by the SR model;
after this, the scheduler feeds the up-scaling frame into the inference DNN for inference (\eg, object detection).
The frames in the inference class enjoy the benefits from the SR frames following the references in Fig. \ref{fig:IIIA}, and then are fed into the inference DNN model for inference.
Their accuracy still increases due to the benefits transferred from the SR frames.
Note that the transfer is quite fast (the time cost is the same as normal frame decoding) as it only includes additional bicubic interpolation on residual per frame compared to normal frame decoding.
For those frames in the reuse class, \eg, object detection, we get the Bbox of each object in the last (playback order) detected frame, calculate the mean of all MVs that reside in the Bbox, and use it to shift the previous position to the current position.

\tb{Model for pipeline selection}.
We formulate the adaptive pipeline selection problem to maximize the accuracy under the latency constraint.
Given a video containing the set of frames $F$, one of the three pipelines is selected for each frame, which can be expressed as follows.
\begin{equation}
\begin{aligned}
\max_{\mathbf{x}} \; & \sum_{f\in F} Acc(x_f) \\
\text{s.t.}  \;\;    & \sum_{f \in F} Latency(x_f) \leq \tau,
\end{aligned}
\label{eq:model}
\end{equation}
where $\mathbf{x}=\{x_1,..., x_F\}$ is the selection set and $x_f\in \{1,2,3\}$ is for pipeline selection, $Acc$ is the accuracy of a frame, $Latency$ is the latency of a frame given the selected pipeline, and $\tau$ is latency tolerant of frames $F$.




\begin{figure}[htp!]
\hspace{-1em}
\centering
\subfigure[Correlations] {
 \label{fig:FvR-a}     
\includegraphics[width=0.238\textwidth]{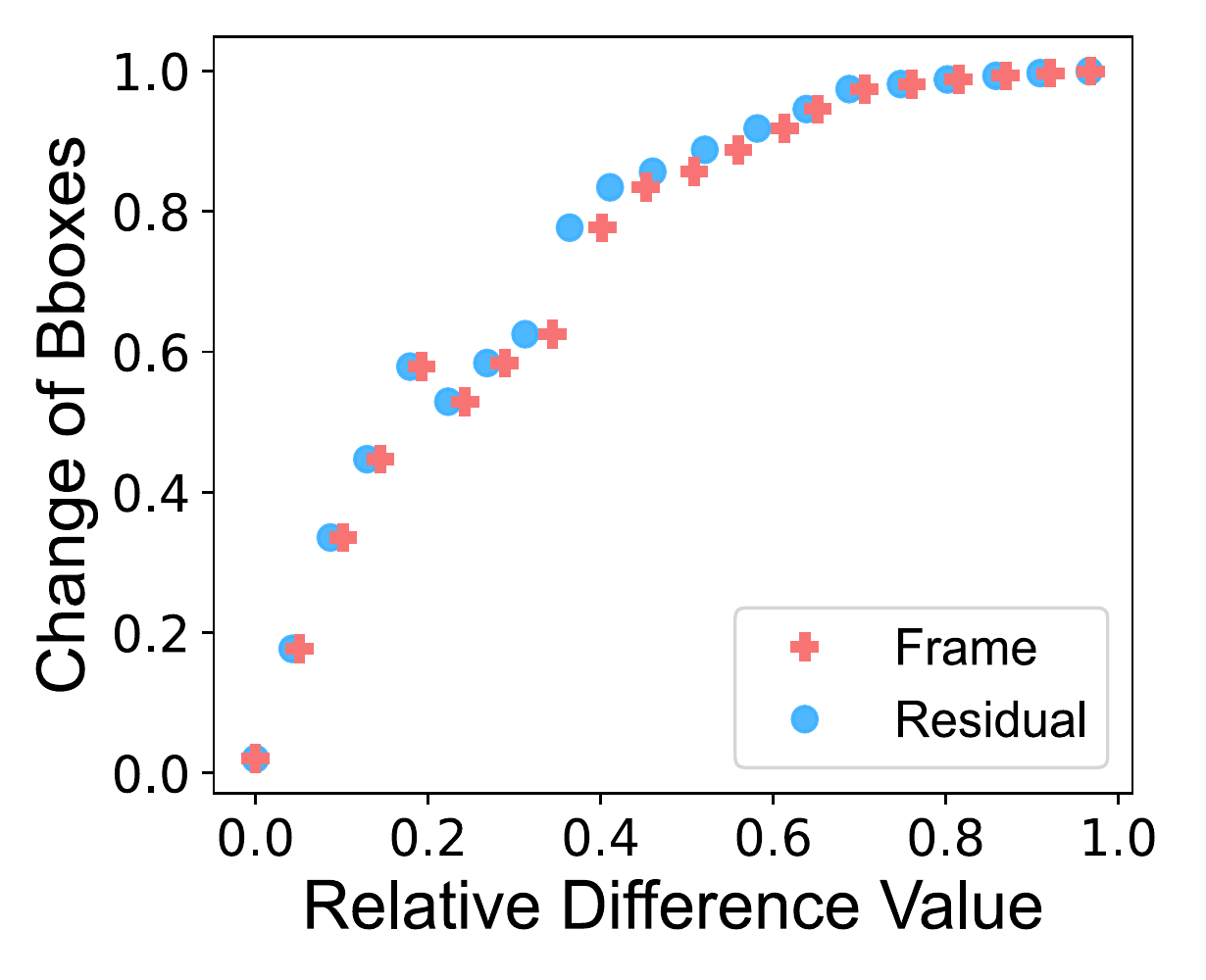}
} 
\hspace{-1.5em}
\subfigure[Time cost] {
 \label{fig:FvR-b}     
\includegraphics[width=0.225\textwidth]{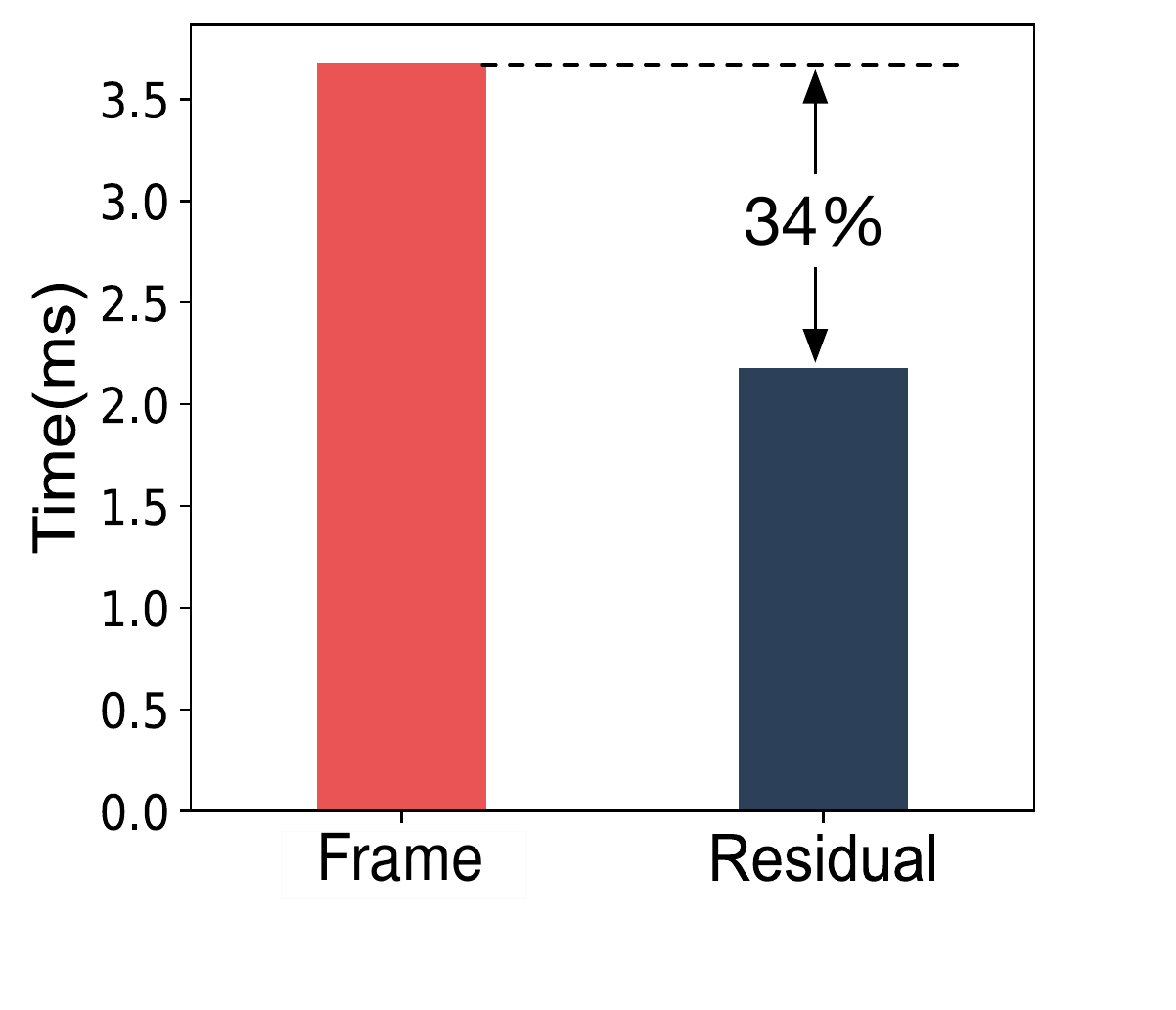}
}
\caption{Frame vs. residual in correlations between the difference values and the changes of Bboxes, and time cost in feature extraction.}
\label{fig:FvR}
\end{figure}

Finding the optimal pipeline selection is tricky due to the large searching space $3^{|F|}$.
Inspired by Reducto \cite{li2020reducto} which adaptively filters frame via setting a threshold on frame differencing,
we introduce two thresholds $tr_1$ and $tr_2$ on frame differencing to cluster frames into three pipelines.
When the frame difference is greater than $tr_1$, the frame will enter the pipeline \ding{182}, and similarly, $tr_2$ is the threshold for the pipeline \ding{183}.
If the difference of a frame is greater than both of them, it will enter the pipeline\ding{182}.
The constraint of real-time video analytics (\eg, speed of analytics $\geq$30fps) restricts us from extracting the light-weighted features to categorize frames.
Different from \cite{li2020reducto}, we find out the Laplacian (\ie, edge features) on the residual and the frames have a high correlation with the inference accuracy (see Fig. \ref{fig:FvR-a}).
At the same time, executing the Laplacian operator on the residual can save 34\% time than that on frame (see Fig. \ref{fig:FvR-b}).
One intuitive reason is that information on residuals is sparse and de-redundant.
It preserves differences among frames but is not too dense to process, thus providing a good opportunity to categorize frames efficiently.


\begin{figure}[htp!]
\centering
\subfigure[Video of crossroad] {
 \label{fig:BT-a}     
\includegraphics[width=0.232\textwidth]{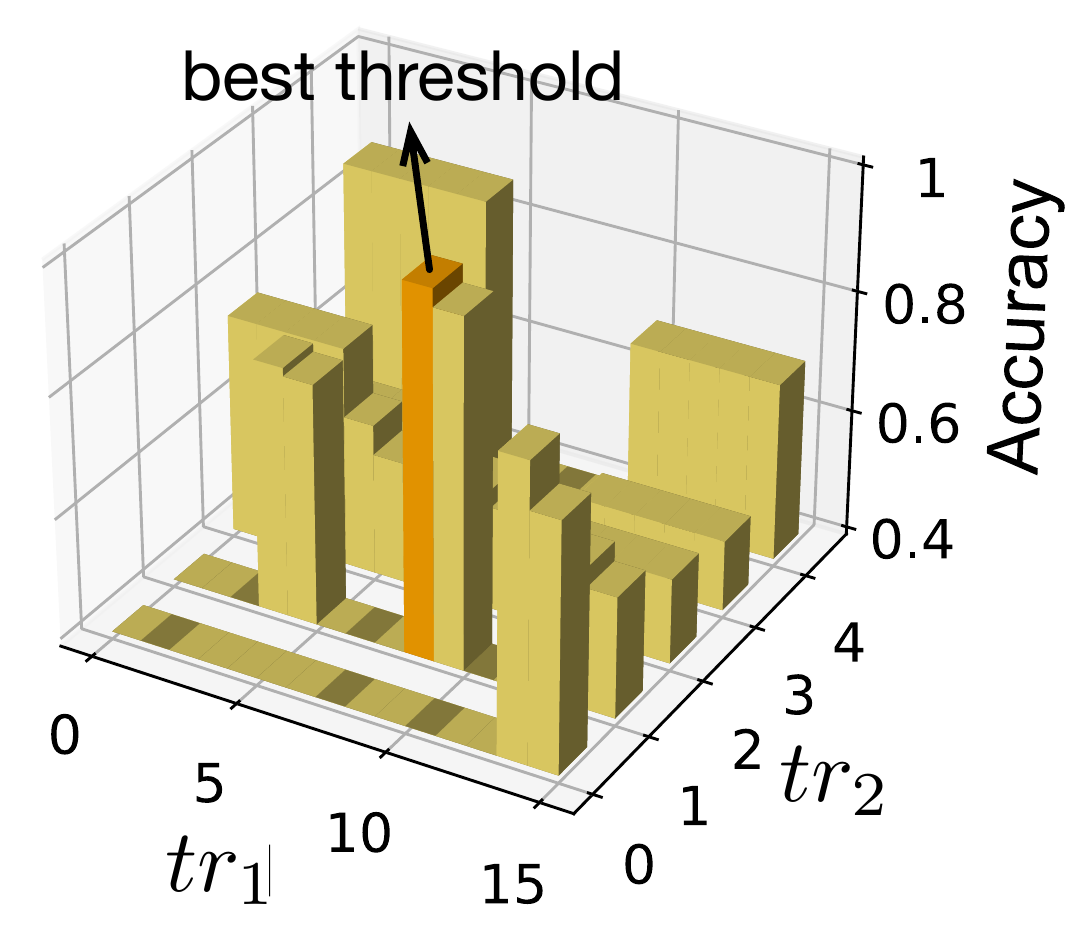}
}
\subfigure[Video of highway] {
 \label{fig:BT-b}
 \hspace{-1em}
\includegraphics[width=0.232\textwidth]{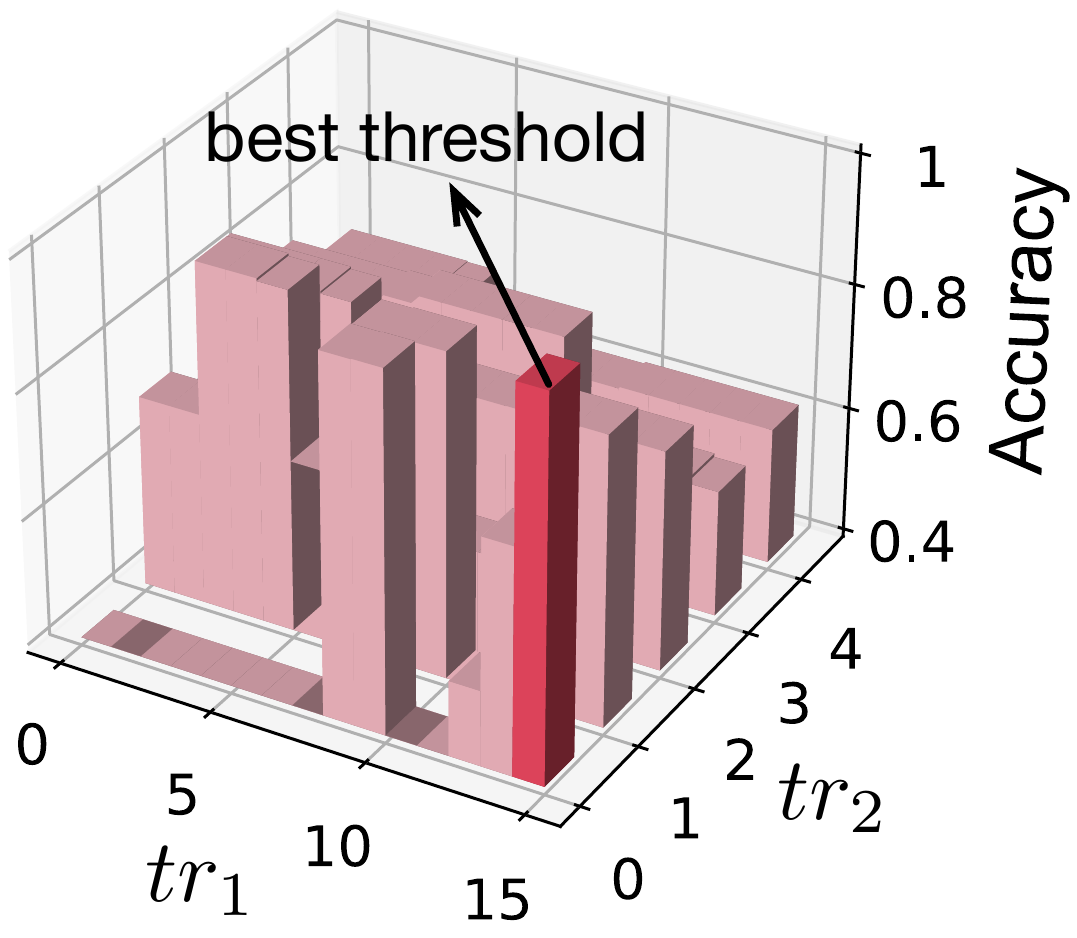}
}
\caption{Best thresholds for pipeline selection vary across videos.}
\label{fig:BT}
\end{figure}

\begin{figure}[h!]
\centering
    \includegraphics[width=0.47\textwidth]{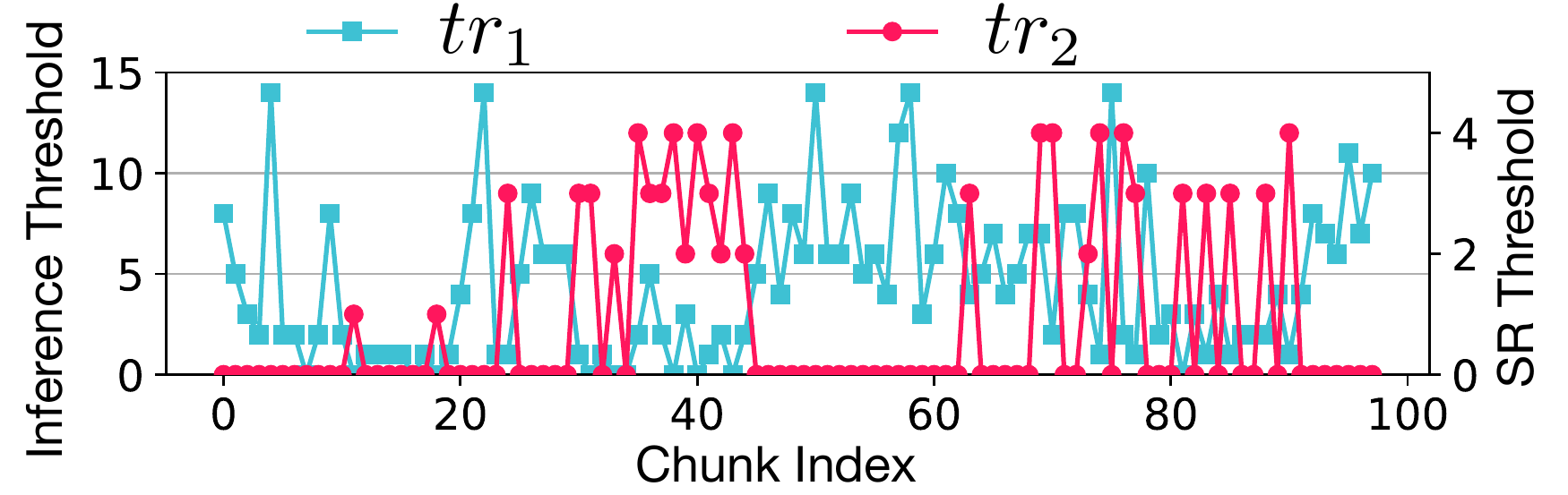}
    \caption{Best thresholds vary across chunks (even adjacent).}
    \label{fig:chunk}
    \vspace{-0.5em}
\end{figure}

\tb{Solution}.
Categorizing frames into three classes for pipelines is not trivial.
Across various videos, their best threshold combination differs (as shown in Fig. \ref{fig:BT}).
Furthermore, the optimal thresholds for frame feature differences (\eg, pixel and residual differences) among chunks in one video vary greatly.
Fig. \ref{fig:chunk} plots the best thresholds on the videos in \cite{Gebhardtdata}, which implies we should dynamically adjust the threshold for each chunk.
Therefore, the scheduler needs to offer adaptive threshold settings.

To adaptively set the thresholds, we formulate this problem as a Markov decision process (MDP) where the scheduler makes the threshold setting decision in AccDecoder.
The MDP is a discrete-time stochastic process, which can be defined by a quad-tuple $<S, A, R, P>$. In this tuple, $S$ is the set of states, $A$ is the set of actions, $R$ is the set of rewards, and $P$ is the probability of transition from state $S$ to state $S'$ based on action $A$.
While processing frames, the scheduler's goal is to cluster them into three pipelines (\ie, the action $A$) to maximize its expected long-run reward $\mathbb{E}[R_i]$.
We define how the MDP is parameterized as follows.
\begin{itemize}[left=0.2em]
    \item \textbf{State}:
    The state consists of two components:
    the content feature of the key frame
    and the differencing features including inter-frame differences and difference between the key frame and the last inference frame.
    First, the features of the key frame (\ie, the first frame of the current chunk) are extracted through the 1x1x1000 fully connected FC-1000 layer of VGG16 \cite{simonyan2014very}.
    Since the dimension of this feature is too large, we use principal component analysis (PCA) to reduce it to 128 dimensions.
    Next, we compute the inter-frame differences between every two frames of each chunk, which is the difference of edge features (\ie, apply the Laplacian operator to the residual of each frame) as discussed in Section \ref{features}.
    Considering that there is continuity between chunks, it is also necessary to add information about inter-chunk, that is, the difference of edge features between the key frame and the last inference frame in the previous chunk.
   
    \item \textbf{Action}: The action is to set two thresholds $tr_1$ and $tr_2$ for each chunk.
    The first threshold $tr_1$ is applied to these frames to select anchor frames for SR and transfer their quality to the others for scale-up.
    The second one $tr_2$ is to select inference frames that need to be analyzed by inference DNN.
    Then, the rest of the frames reuses the inference results by exploring frame reference.
    It is challenge to make accurate decisions for a large space of actions \cite{yan2021acc}. Therefore, we discretize the action space to reduce the search space, \ie, $tr_1 \in \{ 0.05,0.10,0.15,...,0.75\}$ and $tr_2 \in \{ 0.5,1.0,...,2.5\}$.

    \item \textbf{Reward}:
  Given that AccDecoder aims to maximize the inference accuracy within a tolerable latency.
  Therefore, the reward is designed to include two aspects, namely, the average accuracy of the chunk and the latency required to obtain the inference results for the chunk.
  Our goal is to achieve real-time inference, so the chunk needs to be analyzed before the arrival of the next chunk (\eg, within 1s for each chunk); otherwise, there is a penalty for exceeding the specified time.
    The reward for each chunk $t$ is defined as follows.
    \begin{equation}
    r_t= \frac{\alpha_1}{|F_t|} \sum_{f\in F_t} Acc_f(tr_{t,1}, tr_{t,2})-\alpha_2 P_t(tr_{t,1}, tr_{t,2}),
    \end{equation}
    where $\alpha_1$ and $\alpha_2$ are the weight factors to balance the preference for latency and accuracy. The value of $\alpha$ in reward can be adjusted based on different service preferences and requirements.
    $P_t$ is a penalty function of chunk $t$ for latency exceeding the tolerance $\tau$.
    \begin{equation}
        P_t(tr_{t,1}, tr_{t,2}) =\begin{cases}
            1& \text{ if } Latency(tr_{t,1}, tr_{t,2}) > \tau, \\ 
             0 & \text{ others.} 
            \end{cases}
    \end{equation}
    
\end{itemize}

The process of DRL in frame selection is shown below.
At each chunk $t$, the agent observes the current state $s_t$ and gives an action $a_t$ according to its policy. Then, the environment returns reward $r_t$ as feedback, and moves to the next state $s_{t+1}$ according to the transition probability $P(s_{t+1}|s_t, a_t)$.
The goal to find an optimal policy can thus be formulated as the mathematical problem of maximizing the expectation of cumulative discounted return $R_{t} = \sum_{k=t}^T\gamma^{k-t}r_{k}$, where $\gamma\in [0,1]$ is a discount factor for future rewards to dampen the effect of future rewards on the action; $r_k$ is the reward of each step, and $T$ is the number of chunks in the video. 


\tb{Computational complexity}.
The searching space of the optimal pipeline selection for a chunk is $\mathcal{O}(3^{k})$, where $k$ is the number of frames in the chunk.
The searching space of AccDecoder's scheduler is $\mathcal{O}(|a|)$, where $|a|$ is the number of possible actions.
As we discretize the action space, therefore, the complexity of AccDecoder is much less than that of optimal pipeline selection.
\section{Implementation}
We implement AccDecoder as an intelligent decoder in both simulation and prototype.
We first introduce the system settings of the experiment and the setup of the dataset and the baselines.
Next, we introduce the training setup of AccDecoder on neural networks of DRL, SR, and inference.

\subsection{System Settings, Dataset, and Baselines}

\tb{System settings}.
The server component runs on an Ubuntu 18.04 instance with Intel(R) Xeon(R) Gold 6226R CPU at 2.90GHz and 1 NVIDIA GeForce RTX 3070 GPU.
AccDecoder is built on H.264 codec \cite{libvpx}, JM 19.0 version open source code \cite{JM}, with 2470 LoC changes.
In practice, there are several video codecs besides H.264, but they have a high degree of similarity.
For example, they share the same abstracts (\ie, reference index, MV, and residual), which are essential information to transfer neural super-resolution outputs to non-anchor frames.
Their difference is in low-level compression algorithms (\eg, the number of reference frames and the size of blocks).
Thus, while we only use H.264 to validate AccDecoder’s design, we believe the design is generic enough to accommodate different codecs.



\tb{Dataset}.
Our video dataset mainly contains public data streams from real-time surveillance cameras deployed worldwide.
We collected video clips of different scenes and times from different camera data sources.
Thus, video datasets with different properties (\eg, time, illumination, vehicle and pedestrian density, road type, and direction) are obtained.
All of our videos are available by searching on YouTube \cite{Gebhardtdata,jackson,auburn2} and Yoda \cite{Yoda_c,Yoda_m,Yoda}.
The videos are in 30fps, of which each chunk includes 30 frames (\ie, $k=30$).
In particular, VisDrone dataset \cite{fan2020visdrone} is used to train SR model.
We use CityScapes \cite{cordts2016cityscapes} as dataset and results of ERFNet \cite{romera2017erfnet} to calculate inference loss for training.

\noindent
\textbf{Baselines}.
We compare AccDecoder’s performance with the following four baselines:
(1) Glimpse \cite{chen2015glimpse} filters frames by comparing pixel-level frame differences against a static threshold.
(2) AWStream \cite{Awstream} adapts encoding parameters (\ie, quantization parameter (QP), resolution, and frame rate) of the underlying codec to cope with different available bandwidth;
(3) DDS \cite{du2020server} applies different quality encodings to different regions via region proposal network (RPN) \cite{ren2015faster}, which can offer trade-off accuracy and latency.
(4) Reducto \cite{li2020reducto} filters unnecessary frames in its encoder via a dynamic setting threshold given by the server to save bandwidth in uploading.

\subsection{DNN Implementation}

\noindent
\textbf{DRL settings}.
In the experiments, we set $\alpha_1=\alpha_2=0.5$ and $\tau=1s$ for each chunk according to our practical experience and requirements of services.
We use an Adam optimizer with a learning rate of 0.0001. The discount factor for reward $gamma$ is 0.99.
We use a two-layer MLP with 128 units to implement the policy networks in DRL.
The neural networks use ReLU as activation functions.
The capacity of the replay buffer is $10^5$, and we take a minibatch of $256$ to update the network parameters.

\noindent
\textbf{SR model}. For object detection, we train the detection-driven SR model based on EDSR \cite{lim2017enhanced} following the analytics aware loss function (\ie, a weighted addition of visual quality loss and object detection inference loss) in \cite{wang2022enabling}.
We use VisDrone dataset \cite{fan2020visdrone} as the training set  to train our model; use testing set from VisDrone, video set from DDS \cite{du2020server} and Reducto \cite{li2020reducto} to evaluate its performance.
The visual quality loss comes from the pair of original and down sampling frames; the object detection inference loss comes from the results of YOLOv4 \cite{bochkovskiy2020yolov4} detecting on reconstructed frames (from the down sampling frames) and the labels.
The initial weights of EDSR and YOLOv4 are provided by authoritative implementations \cite{Lim_2017_CVPR_Workshops, bochkovskiy2020yolov4}. The weights of EDSR are updated during our training, but the one of YOLOv4 keeps static. 


\noindent
\textbf{Inference DNN settings}.
We mainly evaluate AccDecoder's performance on object detection.
Here, we list the different DNNs used by AccDecoder for object detection. We choose two object detection models with different architectures including Faster R-CNN \cite{ren2015faster}, YoLov5 \cite{yolo}. We pre-trained both models using the COCO dataset \cite{COCO}.

\section{Evaluation}
To demonstrate AccDecoder delivers significant quality improvement, we compare AccDecoder with baselines in terms of inference accuracy and latency.


\subsection{Overall performance of AccDecoder}

\begin{figure}[h!]
\centering
    \subfigure[Ground truth (1080p)] {
     \label{fig:SR_frame-a}     
    \includegraphics[width=0.37\textwidth]{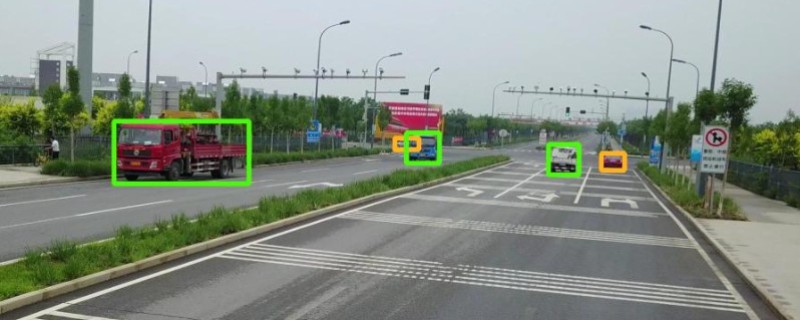}
    }
    \hspace{-1.3em}
    \subfigure[Low resolution (270p)] {
     \label{fig:SR_frame-b}     
    \includegraphics[width=0.37\textwidth]{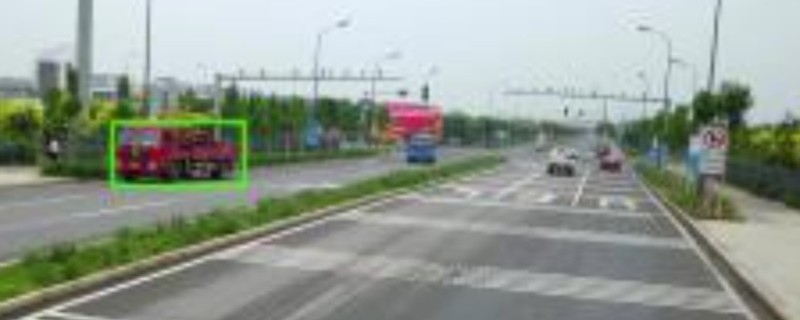}
    }\\ 
    \subfigure[Super resolution]{
     \label{fig:SR_frame-d}     
    \includegraphics[width=0.37\textwidth]{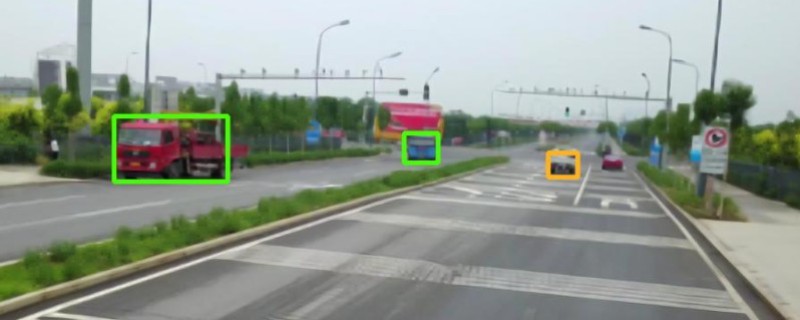}
    } 
    \caption{SR can improve inference accuracy by enhancing resolution of frames in videos.}
    \label{fig:SR_frame}
\end{figure}

\noindent
\tb{Video quality improvement}.
We illustrate the performance of SR via Fig. \ref{fig:SR_frame}, which shows the visual object detection results.
The results vary in the original 1080p images with ground truth labels, inference results of low resolution (in 270p), and inference results of up-scaled images by SR from 270p.
From the results, we can see SR delivers more detailed information to the DNN inference model, thus bounding more small objects and improving accuracy in object detection. 

\begin{figure}[h!]
\centering
    \includegraphics[width=0.48\textwidth]{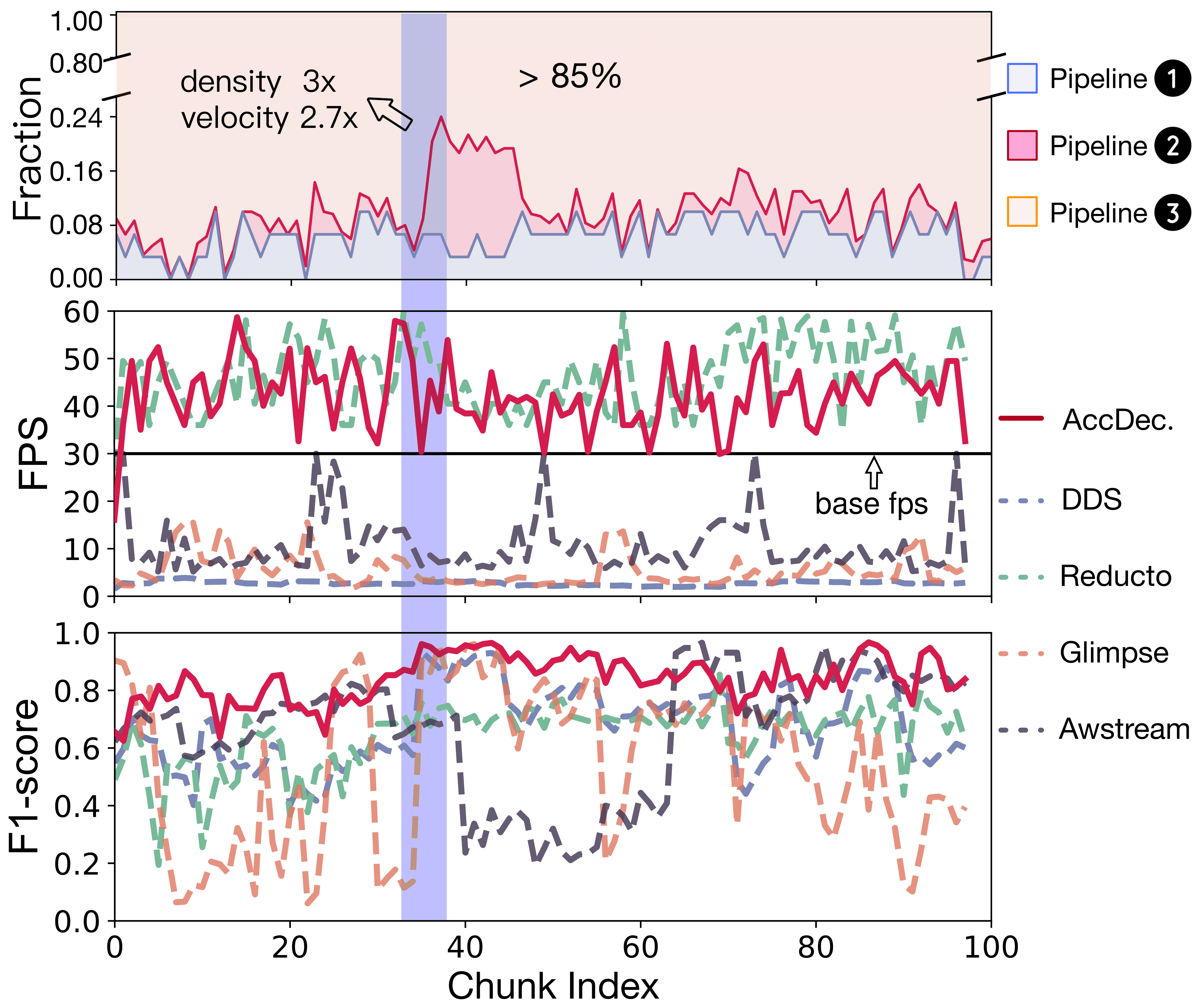}
    \caption{Pipeline assignment ratio, analytics speed, and inference accuracy of chunks.}
    \label{fig:chunks}
\end{figure}

\tb{Accuracy and latency}.
We demonstrate that AccDecoder can effectively improve accuracy-latency trade-off via adaptive pipeline assignment.
Fig. \ref{fig:chunks} shows the per-chunk of pipeline assignment, accuracy  (\ie, f1-score as a measure of accuracy), and speed of analytics, respectively.
AccDecoder clusters frames into three types:
1) anchor frames (around 6\%) are selected for pipeline \ding{182} with SR;
2) inference frames (11\%) are analyzed by inference DNNs in pipeline \ding{182} and \ding{183}, in which 5\% frames are for pipeline \ding{183};
3) non-inference and non-anchor frames (around 89\%) are selected for pipeline \ding{184}.
Furthermore, AccDecoder and Reducto can achieve a higher frame rate than the basic requirement (\ie, 30fps), whereas other baselines offer a lower rate.
AccDecoder can also achieve higher and more stable accuracy than the baselines.
It is worth noting that from the 35th chunk, the density and velocity of vehicles increase significantly (\ie, 3x and 2.7x, respectively) so that the inferred rate is adjusted to a higher level, which ultimately maintains a good accuracy and frame rate.

\begin{figure}[h!]
\centering
\vspace{-1em}
\hspace{-4em}
\subfigure[Accuracy breakdown] {
 \label{fig:break-a}     
\includegraphics[width=0.3\textwidth]{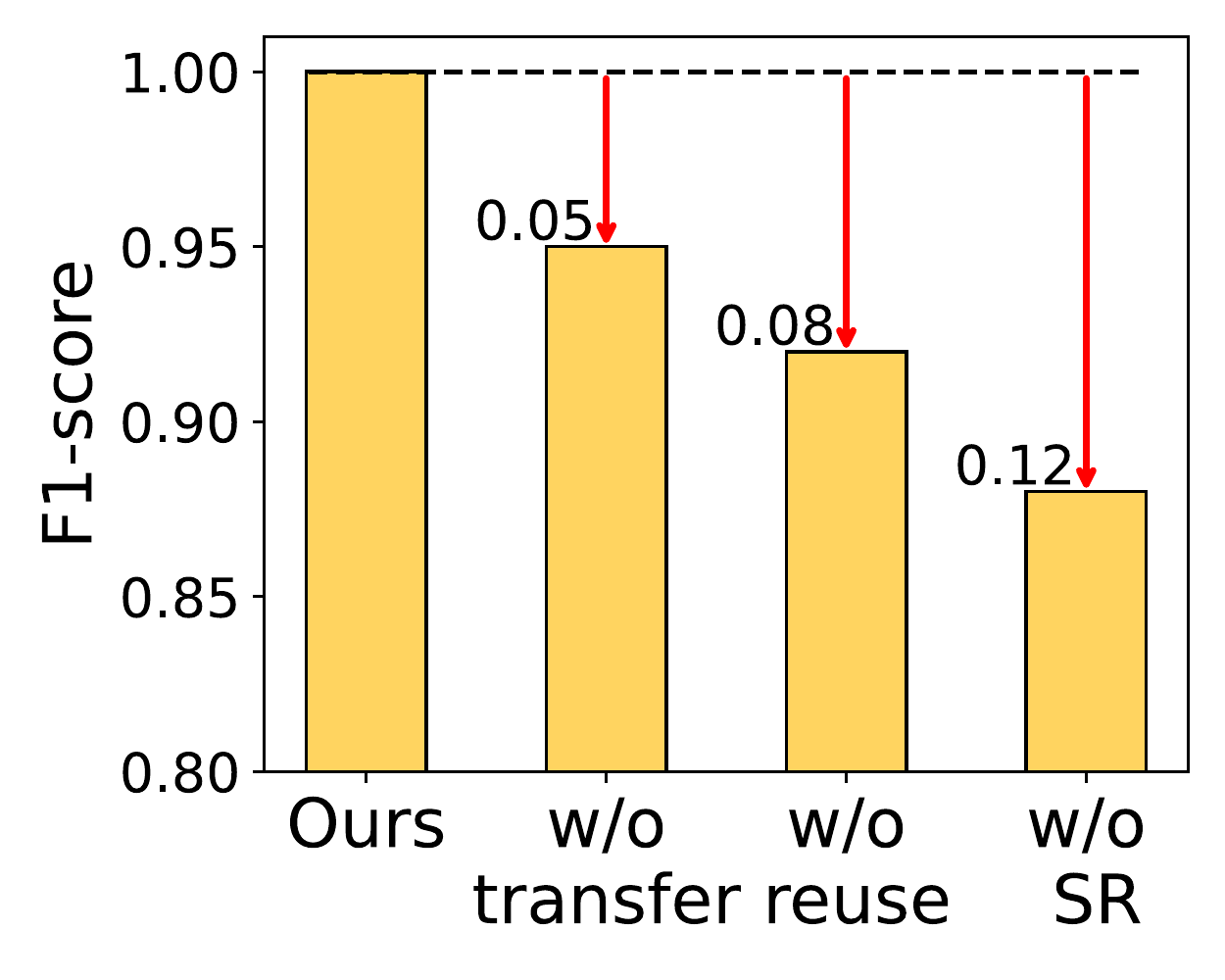}
}
\hspace{1.5em}
\subfigure[Latency breakdown] {
 \label{fig:break-b}
\includegraphics[width=0.42\textwidth]{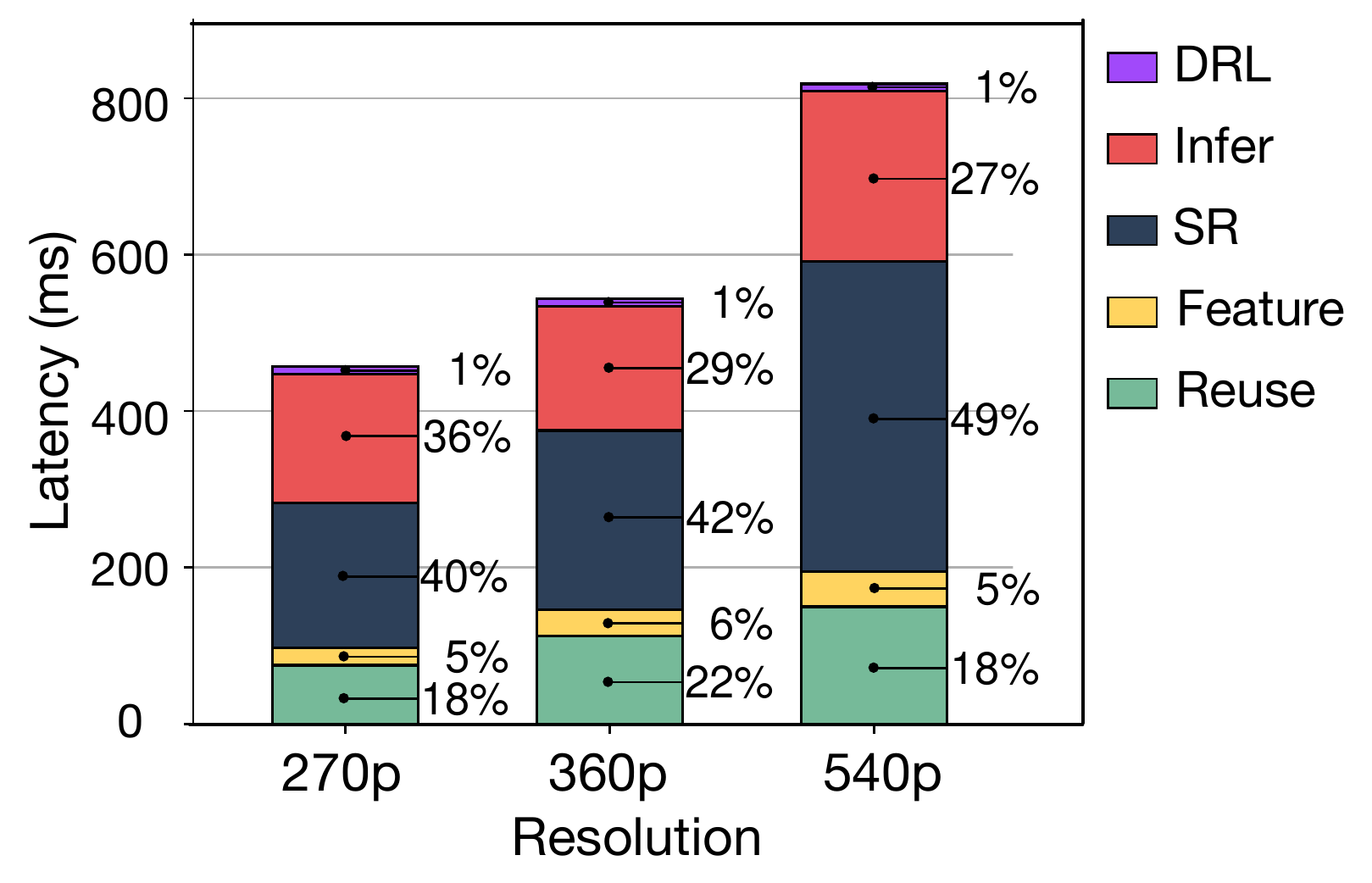}
}
\vspace{-0.5em}
\caption{Performance breakdown on accuracy and latency.}
\label{fig:break}
\end{figure}

 
\subsection{Component-wise Analysis}

\tb{Performance breakdown}.
We break down the accuracy and the latency to show the impact of each pipeline illustrated in Fig. \ref{fig:break}.
Firstly, Fig. \ref{fig:break-a} breakdowns the accuracy in transfer of anchor frames, reuse of the inference frames, and SR.
The impact of reuse and SR on accuracy is obvious, while the impact of transfer is relatively small.
The main reason is that the transfer has less improvement on small objects due to blur caused by interpolation, which will be one of our future works.
Secondly, Fig. \ref{fig:break-b} shows the latency breakdown in processing each chunk, where we use videos in 270p, 360p, and 540p, respectively.
AccDecoder needs more time when processing and analyzing high-resolution videos, \ie, the overall latency in ms.
Specifically, SR ($\sim$40\%), inference ($\sim$30\%), and reuse ($\sim$20\%) take up most of the latency, while DRL-based scheduling and feature extraction only occupy around 5\%, which can be omitted.
SR is time-consuming, although only 6\% of frames are assigned to pipeline \ding{182}.
Besides, inferring pipeline \ding{182} and \ding{183} for around 11\% of frames also takes a latency that cannot be ignored.
Although the time cost of reuse per frame is small, the time cost of reuse requires 20\% latency in total due to more than 85\% frames belonging to pipeline \ding{184}.

\begin{figure}[h!]
\centering
    \includegraphics[width=0.49\textwidth]{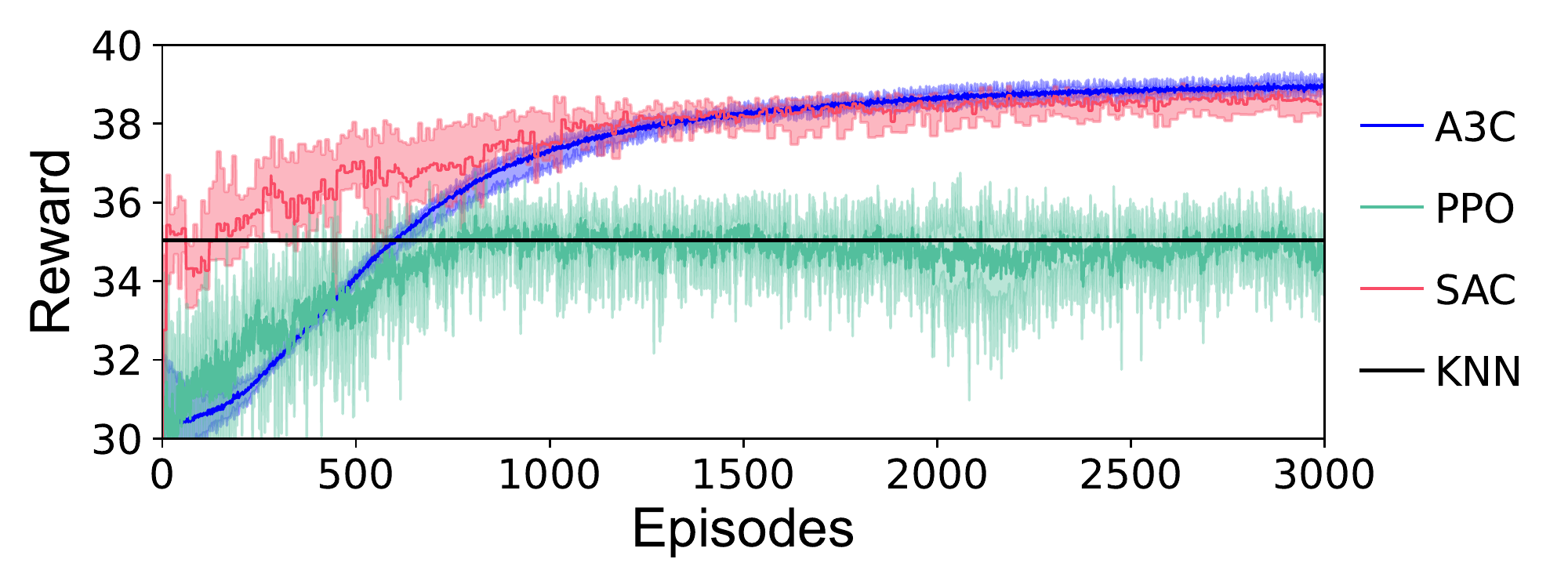}
    \caption{Comparison of different schemes for the scheduler in AccDecoder. DRL is better than KNN, and A3C achieves better learning performance than the others.}
    \label{fig:drl}
\end{figure}

\vspace{1em}
\tb{AccDecoder schedulers}.
We evaluate the scheduler's performance in AccDecoder, including DRL-based schemes and KNN, as illustrated in Fig. \ref{fig:drl}.
We choose three well-known DRL schemes for training: Asynchronous Advantage Actor-Critic (A3C)~\cite{mnih2016asynchronous}, Soft Actor-Critic (SAC)~\cite{haarnoja2018soft}, and Proximal Policy Optimization (PPO)~\cite{schulman2017proximal}.
Through effective exploration, we find A3C can receive a satisfactory and stable cumulative reward, which was about 11.42\% higher than the cumulative reward received by KNN and PPO.
From the final cumulative reward value in the figure, the cumulative rewards of A3C and SAC are higher. Considering the advantages of A3C, we choose A3C for the training of the scheduler.

\subsection{AccDecoder vs. Existing VAPs}

\begin{figure}[htp!]
\centering
\hspace{-1em}
    \subfigure[Faster R-CNN (highways)] {
     \label{fig:bub-a}     
    \includegraphics[width=0.235\textwidth]{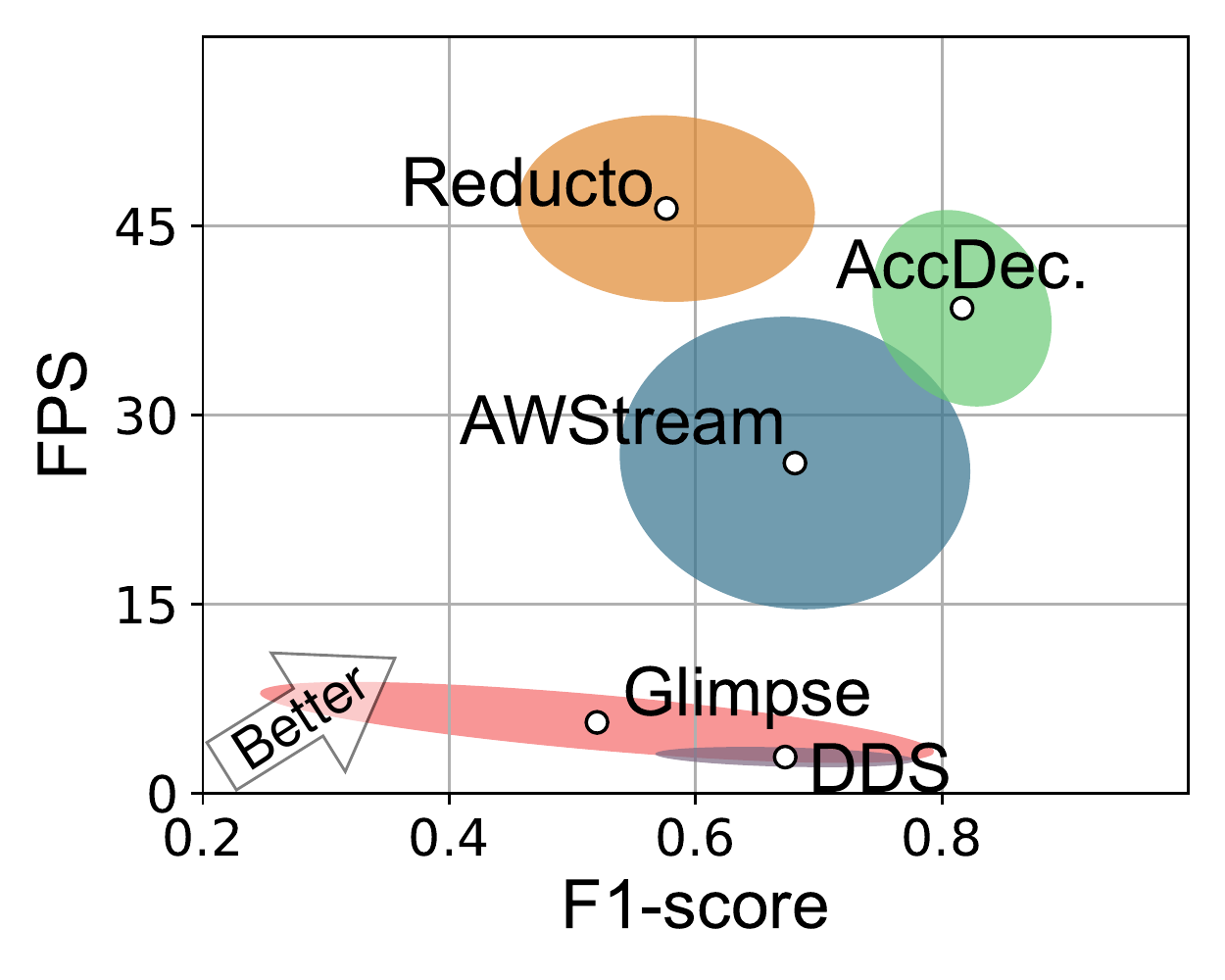}
    }
    \hspace{-1em}
    \subfigure[Yolov5 (highways)] {
     \label{fig:bub-b}     
    \includegraphics[width=0.235\textwidth]{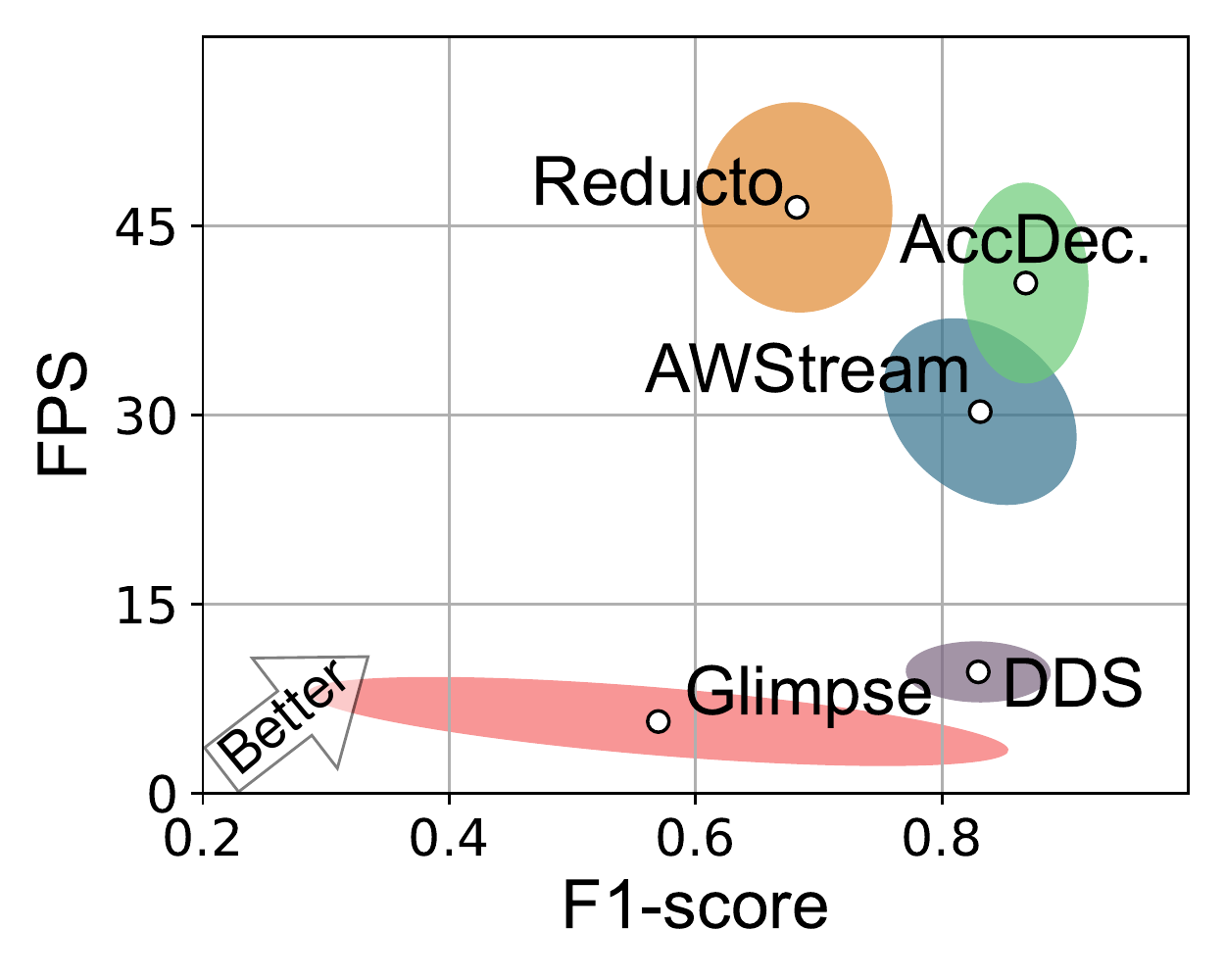}
    } \\
    \hspace{-1em}
    \hspace{-1em}
    \subfigure[Faster R-CNN (crossroad)] {
     \label{fig:bub-c}     
    \includegraphics[width=0.235\textwidth]{fastcro.pdf}
    } 
    \hspace{-1em}
    \subfigure[Yolov5 (crossroad)] {
     \label{fig:bub-d}     
    \includegraphics[width=0.235\textwidth]{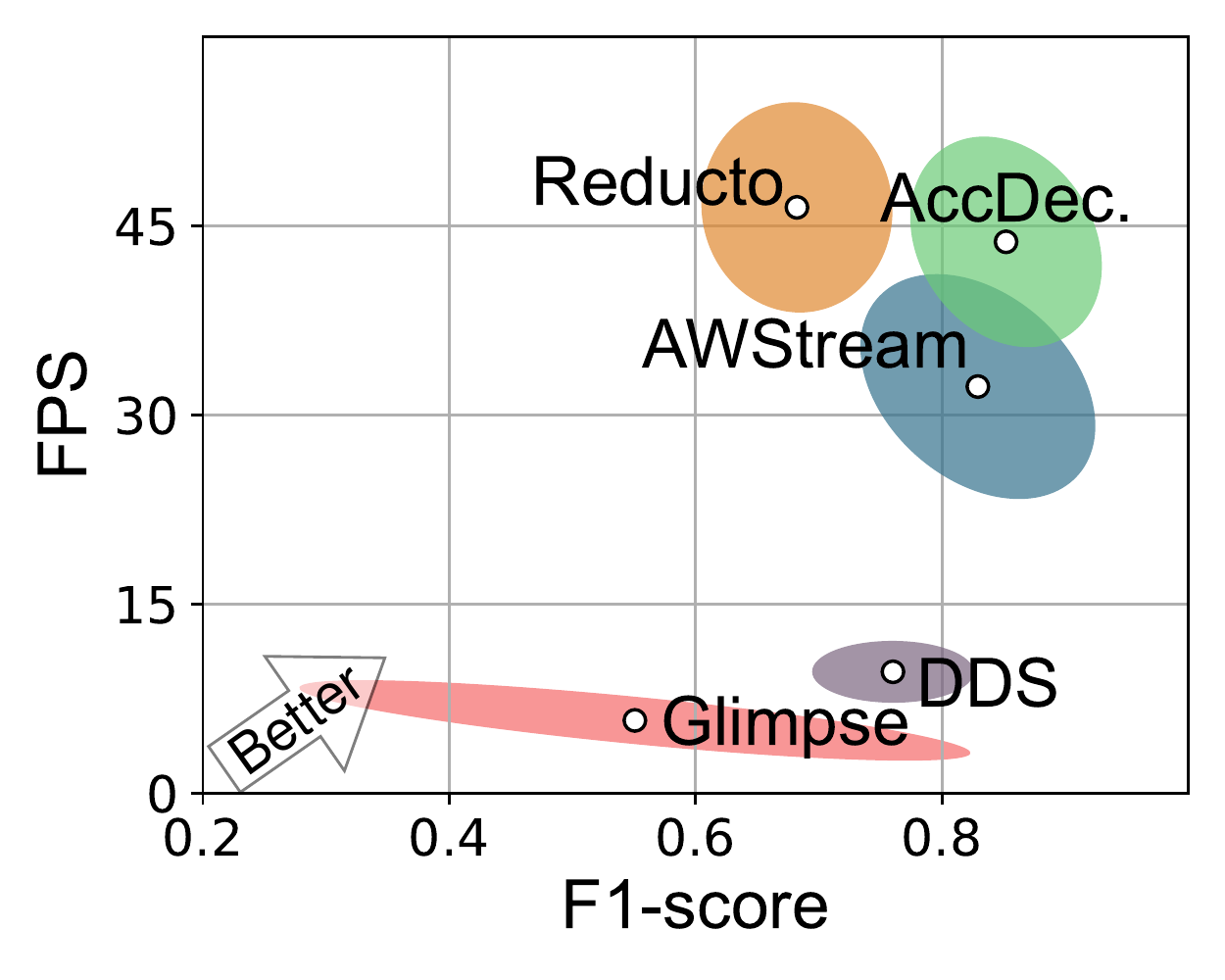}
    }
    \hspace{-1em}
    \caption{AccDecoder vs. baselines in terms of the speed of analytics and inference accuracy on various video datasets (in parentheses) and different DNN models (\ie, Faster R-CNN and Yolov5).
    AccDecoder achieves 6-38\% higher inference accuracy than the baselines and 20-80\% lower latency than the baselines except for Reducto.}
    \label{fig:bubbles}
    \vspace{-0.5em}
\end{figure}

\tb{AccDecoder vs. baselines}.
We show the advantage of AccDecoder in accuracy and speed of analytics compared with the baselines.
Fig. \ref{fig:bubbles} compares the performance distribution of AccDecoder with the baseline over different DNNs (YOLOv5 and Faster R-CNN with Resnet50) and various types of videos (\ie, highways and crossroads).
It can be seen that AccDecoder is better than the baseline in terms of speed and accuracy of analytics.
In terms of speed, AccDecoder's analytics speed is around 35fps, which meets the penalty threshold we set for DRL training.
In terms of accuracy, AccDecoder can keep relatively higher accuracy compared with the baselines.

\section{Related Works}

\noindent
\textbf{Video analytics pipeline}.
Many computer vision tasks are considered in VAPs, such as traffic control \cite{ananthanarayanan2017real}, surveillance and security \cite{yi2020eagleeye}.
We consider the following task as a running example --- object detection.
Object detection aims to identify objects of interest (\ie, their locations and classes) in each frame in the video.
Selecting this task has two major reasons: first, it plays a core role in the computer vision community because a wide range of high-level tasks (\eg, autonomous driving) is built on it;
second, we seek to keep consistent with prior video analytics work \cite{du2020server,jiang2018chameleon,jain2020spatula,Yuan_VSiM_2022} to allow a straightforward performance comparison.


\tb{Deep reinforcement learning}.
DRL is well suited to tackle problems requiring longer-term planning using high-dimensional observations, which is the case of dynamic pipeline selection.
There is a variety of DRL-based algorithms. Value-based algorithms, \eg, Deep Q-Networks (DQN)~\cite{mnih2015human}, use a deep neural network to learn the action-value function. However, they do not support continuous action space like the one in our problem. Policy-based algorithms, \eg, Policy Gradient (PG)~\cite{sutton1999policy}, explicitly build a representation of a policy. However, evaluating a policy without action-value estimation is typically inefficient and causes high variance. Actor-critic algorithms learn the value function (critic) in addition to the policy (actor) since knowing the value function can assist policy updates, for example, by reducing variance in policy gradients. Many existing approaches are based on actor-critic, for example, PPO~\cite{schulman2017proximal}, A3C~\cite{mnih2016asynchronous}, and SAC~\cite{haarnoja2018soft}.
A3C, an asynchronous algorithm, can enable multiple worker agents to train in parallel, allowing faster training.

\section{Conclusion and Discussion}
In this paper, we propose AccDecoder to eliminate the dependence of existing VAPs on video quality.
AccDecoder is a new universal video stream decoder that uses a super-resolution deep neural network to enhance video quality for video analytics.
To accelerate analytics, AccDecoder applies DRL for adaptive frame selection for quality enhancement or/and DNN-based inference.
It is a new way to address the key challenge of accuracy-latency tradeoff in distributed VAPs.
We show that AccDecoder can substantially improve state-of-the-art VAPs by speeding up analytics (3-7x) and achieving accuracy improvements (6-21\%).

In future work, we plan to explore DRL for macroblock selection to offer finer-grained scheduling and joint adaptation encoding and decoding to further improve the accuracy and speed of analytics.

\section*{Acknowledgment}

This work has been partly funded by EU H2020 COSAFE (Grant 824019) and Horizon CODECO projects (Grant 101092696), the Alexander von Humboldt Foundation, and the National Natural Science Foundation of China under Grant 61872178, 62272223, and 
Grant 61832005.

\bibliographystyle{IEEEtran}
\bibliography{ref}

\end{document}